\documentclass[conference]{IEEEtran}
\IEEEoverridecommandlockouts

\usepackage{nopageno}

\usepackage{amsmath,amsfonts}
\usepackage{algorithmic}
\usepackage{graphicx}
\usepackage{textcomp}
\usepackage{xcolor}
\usepackage{algorithm}
\usepackage[subrefformat=parens,labelformat=parens]{subfig}

\usepackage{cite}  
\usepackage{url}
\usepackage{amsthm}
\usepackage[subrefformat=parens,labelformat=parens]{subfig}
\usepackage{etoolbox}
\newtheorem{theorem}{Theorem}

\newtheorem{lemma}{Lemma}

\usepackage{mathtools}

\usepackage{tikz}
\usetikzlibrary{arrows, positioning, automata}
\usepackage{color,soul}
\usepackage{paralist}
\usepackage{multirow}
\usepackage{graphbox}
\usepackage{epstopdf}
\usepackage[normalem]{ulem}

\usepackage{enumitem}
\setlist[itemize]{align=parleft,left=0pt..1em}

\def\BibTeX{{\rm B\kern-.05em{\sc i\kern-.025em b}\kern-.08em
   T\kern-.1667em\lower.7ex\hbox{E}\kern-.125emX}}

\begin{document}

\title{PAAM: A Framework for Coordinated and Priority-Driven Accelerator Management in ROS 2}
\author{\IEEEauthorblockN{Daniel Enright\IEEEauthorrefmark{1}, Yecheng Xiang\IEEEauthorrefmark{1}, Hyunjong Choi\IEEEauthorrefmark{2}, Hyoseung Kim\IEEEauthorrefmark{1}}
    \IEEEauthorblockA{\IEEEauthorrefmark{1} University of California, Riverside
    \\\{denri006, yxian013,  hyoseung\}@ucr.edu}
    \IEEEauthorblockA{\IEEEauthorrefmark{2} San Diego State University
    \\\{hyunjong.choi\}@sdsu.edu}
}
\maketitle

\begin{abstract}
This paper proposes a Priority-driven Accelerator Access Management (PAAM) framework for multi-process robotic applications built on top of the Robot Operating System (ROS) 2 middleware platform. The framework addresses the issue of predictable execution of time- and safety-critical callback chains that require hardware accelerators such as GPUs and TPUs. PAAM provides a standalone ROS executor that acts as an accelerator resource server, arbitrating accelerator access requests from all other callbacks at the application layer. This approach enables coordinated and priority-driven accelerator access management in multi-process robotic systems. The framework design is directly applicable to all types of accelerators and enables granular control over how specific chains access accelerators, making it possible to achieve predictable real-time support for accelerators used by safety-critical callback chains without making changes to underlying accelerator device drivers. The paper shows that PAAM also offers a theoretical analysis that can upper bound the worst-case response time of safety-critical callback chains that necessitate accelerator access. This paper also demonstrates that complex robotic systems with extensive accelerator usage that are integrated with PAAM may achieve up to a 91\% reduction in end-to-end response time of their critical callback chains.
\end{abstract}


\section{Introduction}\label{sec:intro}
The Robot Operating System (ROS) is a popular middleware platform for robotic applications that supports modular integration of software. With the development of its second version, ROS 2, people's perception of this platform has changed that it could be used for serious industrial robotic applications beyond toy examples and lab experiments. For example, many autonomous driving (AD) companies have adopted Autoware~\cite{kato2018autoware} that is an open-source AD stack built upon ROS. NASA has recently announced its plan to utilize ROS in space robotics and has partnered with Open Robotics and Blue Origin to develop the Space ROS framework that extends ROS 2 with verification and validation requirements~\cite{web:spaceROS}. 

To keep up with these needs, the real-time systems community has recently published several pioneering papers~\cite{casini2019response,tang2020response,blass2021ros,blass2021automatic,choi2021picas,sobhani2023,arafat2022response}. In ROS, applications can be composed of {\em processing chains} of callbacks, which are executed by executors. Executors are processes of the underlying OS and communicate through Data Distribution Service (DDS); hence, ROS can be said to offer a {\em multi-process execution model}. 
Prior studies have successfully analyzed the worst-case end-to-end response time of a callback chain and improved real-time performance and predictability of ROS-based applications, but with the assumption that these chains use only CPU resources.

The high computational demands of modern robotic applications necessitate the use of hardware accelerators, including GPUs, TPUs, and FPGAs. However, such a prolific exploitation of accelerator resources poses several issues to guaranteeing timely execution of safety-critical chains. Current ROS-based applications such as Autoware utilize an unmanaged model for accelerator access, resulting in {\em direct hardware resource invocation}. While this may be acceptable for some applications, it leads to several challenges in real-time predictability (see Sec.~\ref{sec:challenges} for more details). Although work has been done to enhance the analyzability and predictability of real-time robotic systems~\cite{casini2019response,tang2020response,blass2021ros,blass2021automatic,choi2021picas}, the introduction of unmanaged access for accelerators compromises their integrity, i.e., analysis results no longer hold, as the execution order is not explicitly regulated by the OS and device drivers. 

This paper presents PAAM, a \uline{P}riority-driven \uline{A}ccelerator \uline{A}ccess \uline{M}anagement
framework for multi-process robotic applications built on ROS 2. In conventional robotic software design, the callbacks of each executor directly invoke accelerators to execute kernels.  Our approach changes this paradigm and offers {\em accelerator access as a service} to robotic applications. PAAM utilizes a standalone ROS 2 executor that acts as an accelerator resource server, handling accelerator access requests from all other callbacks at the application layer.  Hence, any callbacks of chains requiring access to a specific accelerator send requests to the PAAM server corresponding to that accelerator, and these requests are scheduled by the PAAM server with an explicit consideration of their chain priorities and the degree of device-level prioritization and concurrency provided by the target accelerator. This enables {\em coordinated and priority-driven} accelerator management in ROS 2 and even allows applications to seamlessly access accelerators on remote machines. 
In addition, theoretical analysis of the worst-case response time of chains that include accelerator access is made feasible with our framework design. 

The goal of the proposed framework is to provide predictable and analyzable real-time support for accelerators used by safety-critical chains, effectively allowing predictable execution to occur without making changes to underlying accelerator device drivers. Handling accelerator scheduling in the ROS 2 application layer rather than leaving it to the OS or accelerator driver enables granular control over how specific chains access accelerators, because chain criticality and execution behavior are best observed and respected at the middleware level. We implement this framework specifically for Nvidia GPUs and Google Coral TPUs, but it is directly applicable to other types of accelerators. We evaluated the performance of complex robotic systems with extensive accelerator usage when interfaced with PAAM. We found PAAM to provide up to a 91\% decrease in the end-to-end response times of their most critical callback chains.

\section{Background and System Model}\label{sec:background}
\begin{figure*}[h]
\centering
	\subfloat{
		\includegraphics[width=.82\linewidth]{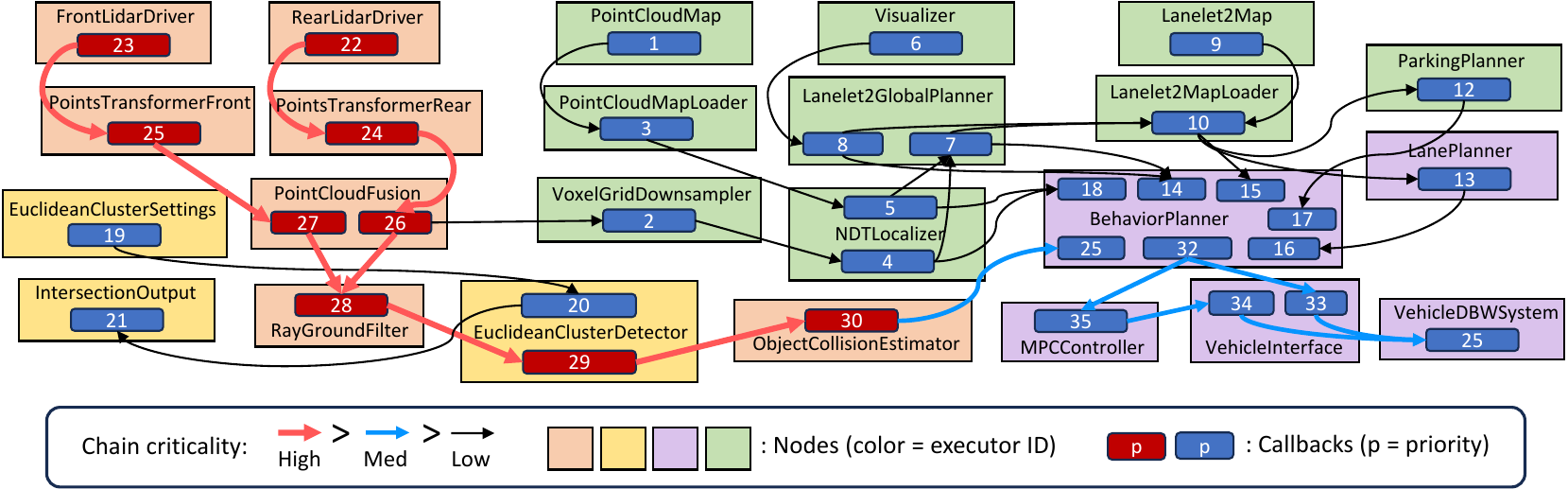}
	}
        \vspace{-1mm}
	\caption{Chain configuration of Apex.AI's Autoware reference system~\cite{web:ros2refsys}: the numbers in rounded boxes are relative callback priorities (higher means higher priority); node colors mean node-to-executor allocation when four single-threaded executors are used; both priority assignment and executor allocation follow the ones provided in \cite{choi2022priority}}
    \label{fig:autoware_model}
\end{figure*}
\subsection{ROS 2 Architecture}
This work considers heterogeneous multi-core platforms with integrated and/or discrete accelerators as well as ROS 2 as a middleware for application development. 
ROS 2 supports a modular integration of software in the form of {\em nodes}. Each node consists of {\em callbacks}, each of which executes as a response to a particular event or message arrival. Callbacks are executed non-preemptively by an {\em executor} that the node has been assigned to. By default, each callback within an executor has a priority that is implicitly determined first by the callback class (i.e., timer, subscription, service, client, waitable) and then within each class by the declaration order in the code~\cite{casini2019response,choi2021picas,tang2020response}. Executors in ROS 2 are OS-level processes and can either be single-threaded or multi-threaded. Callbacks from one or more nodes can be executed by each executor process, depending on the nodal implementation. From the executor's point of view, it does not matter which node callbacks came from, so for ease of presentation, we will not explicitly mention the nodes unless necessary. 

Our PAAM framework is designed to work with standard ROS 2 abstractions such as nodes as standalone modules with sets of callbacks assigned to single-threaded or multi-threaded executors.
Accelerators for this framework can be local or remote execution units, e.g., GPUs, NPUs, and FPGAs, which provide APIs for direct invocation. In this paper, we specifically focus our implementation of this framework on Nvidia GPUs with CUDA workloads and Google Coral Edge TPUs with inferencing workloads.  

\subsection{Processing Chains}
We consider applications as a set of {\em processing chains} of callbacks that must interact in a specific order to complete their missions. For example, in an autonomous vehicle's software stack, a LiDAR-based perception pipeline application may have several chains of data-dependant callbacks that filter ground points, cluster remaining points into detectable objects, filter objects by regions of interest, feature detect on the remaining objects, and finally track objects over time. 
Fig.~\ref{fig:autoware_model} shows the chain configuration of the Autoware reference system developed by Apex.AI~\cite{web:ros2refsys}.
Multiple chains from various nodes are partitioned into four executors, indicated by different node box colors in the figure). Chains have high to low criticality, and any callback from any chain may necessitate access to accelerator resources.


As shown above, each chain at the application level has a criticality although it is not explicitly considered or represented in the standard ROS 2 framework. Recall each callback has a priority implicitly assigned within its executor and each executor is a process scheduled by the OS scheduler. These callback and executor priorities at different layers of the software stack are not aligned with their chain criticalities, and the lack of awareness of chain criticality in ROS 2 is coupled with the fairness-oriented callback scheduling policy of the executor~\cite{casini2019response,blass2021ros,tang2021response}.
These on the one hand help ensure starvation freedom~\cite{blass2021ros}, but on the other hand, make the execution of critical chains unnecessarily delayed by that of non-critical ones~\cite{choi2021picas}. 

To address this issue, PiCAS~\cite{choi2021picas} introduces a mechanism that assigns callback priorities explicitly based on their respective chain criticalities, following the {\em criticality-as-priority (CAPA)} assignment~\cite{de2009scheduling}, and schedules callbacks strictly based on their assigned priorities. We adopt this mechanism in this paper as a starting point due to its superior performance over the vanilla ROS 2. However, as discussed later in Sec.~\ref{sec:challenges}, the use of accelerators brings several new challenges that cannot be solved by PiCAS and other existing solutions. 

\subsection{System Model}
\label{sec:sys_model}


Fundamental ROS 2 abstractions utilized by our framework include executors, callbacks, and chains. We specify related properties such as periods, deadlines, and priorities. 

\smallskip\noindent\textbf{Callbacks.} Callbacks are the smallest schedulable entities in the ROS 2 middleware. 
Each of these callbacks is triggered by some local event such as a timer or incoming message. 

The execution of a callback $\tau_i$ can be viewed as an alternating sequence of CPU and accelerator execution segments. 
\begin{displaymath}
 \tau_i := (E_i, A_i, r_i, \eta_i)
\end{displaymath}
\begin{itemize}[leftmargin=*]
    \item $E_i$: The worst-case execution time (WCET) of CPU segments of one instance (job) of $\tau_i$.
    \item $A_i$: The WCET of accelerator segments of a job of $\tau_i$, i.e., $A_i=\sum {A_{i,j}}$ where $A_{i,j}$ is the WCET of the $j$-th accelerator segment of $\tau_i$.
    \item $r_i$: The set of accelerators used by $\tau_i$, i.e., $r_i=\bigcup r_{i,j}$ where $r_{i,j}$ denotes the accelerator used by $A_{i,j}$. 
    \item $\eta_i$: The number of accelerator segments in a job of $\tau_i$.
\end{itemize}
The WCET of CPU and accelerator segments, $E_i$ and $A_i$, represent the execution time of $\tau_i$ when there are no other interfering tasks on the CPU and accelerators.
If a callback $\tau_i$ does not use any accelerator, $A_i=0$, $r_i=\emptyset$, and $\eta_i = 0$.   


\smallskip\noindent\textbf{Processing Chains.} 
We consider the processing chain model widely used in recent real-time ROS 2 studies~\cite{choi2021picas,casini2019response,tang2020response,sobhani2023}. Specifically, a chain $\Gamma_c$ is characterized as follows:
\begin{displaymath}
\Gamma_c := ([\tau_{c_1}, \tau_{c_2}, ..., \tau_{c_n}], T_c, D_c, \delta_c)
\end{displaymath}
\begin{itemize}[leftmargin=*]
    \item $[\tau_{c1}, \tau_{c2}, ..., \tau_{c_n}]$: The sequence of callbacks executed by each instance of a chain $\Gamma_c$.     
     As in prior work~\cite{choi2021picas,casini2019response,tang2020response,sobhani2023}, we assume that $\tau_{c_{i+1}}$ can start execution only when its predecessor $\tau_{c_{i}}$ completes.
    \item $T_c$: The period of a chain $\Gamma_c$. This is determined by the first callback $\tau_{c_1}$, which is triggered by either a timer or an incoming message.
    \item $D_i$: The relative deadline of $\Gamma_c$. We assume chains have constrained deadlines ($D_c \le T_c$) for admission control.
    \item $\delta_c$: The total number of accelerator segments in each instance of $\Gamma_c$, i.e., $\delta_c =\sum_{\tau_i\in \Gamma_c} \eta_i$.
\end{itemize}
In this model, 
as discussed in prior work~\cite{casini2019response,choi2021picas}, callbacks can be shared among multiple chains and an application can be decomposed into multiple linear processing chains to analyze the end-to-end response time of individual execution flow.

\smallskip\noindent\textbf{Executors.} Executors are a ROS abstraction that serves as a schedulable entity at the OS level. 
In this work, we use the priority-driven callback scheduling mechanism of \cite{choi2021picas}. Hence, an executor schedules its callbacks based on their assigned callback priorities, but just like the default ROS 2 executor, it is non-preemptive. Executors are scheduled on CPU cores by the OS scheduler based on their process priority.

\smallskip\noindent\textbf{Priority Assignment.}
We employ the criticality-as-priority assignment (CAPA) scheme~\cite{de2009scheduling}. 
Hence, chain priorities align with their criticalities, i.e., chains with higher priorities are more critical, and callbacks from a higher-criticality chain have higher callback priorities than those from a lower-criticality one. 
Priority assignments for callbacks, chains, and executors remain static at runtime. The function $\pi()$ indicates the priority of a given entity, e.g., $\pi(\tau_i)$, $\pi(\Gamma_c)$, and $\pi(e_k)$ for callback, chain, and executor priorities, respectively.
Every entity has a unique priority with an arbitrary tie-breaking rule, e.g., if $\Gamma_c\ne \Gamma_c'$, $\pi(\Gamma_c)\ne \pi(\Gamma_c')$.

\begin{figure*}[t!]
\centering
\subfloat{
        \includegraphics[width=.8\linewidth]{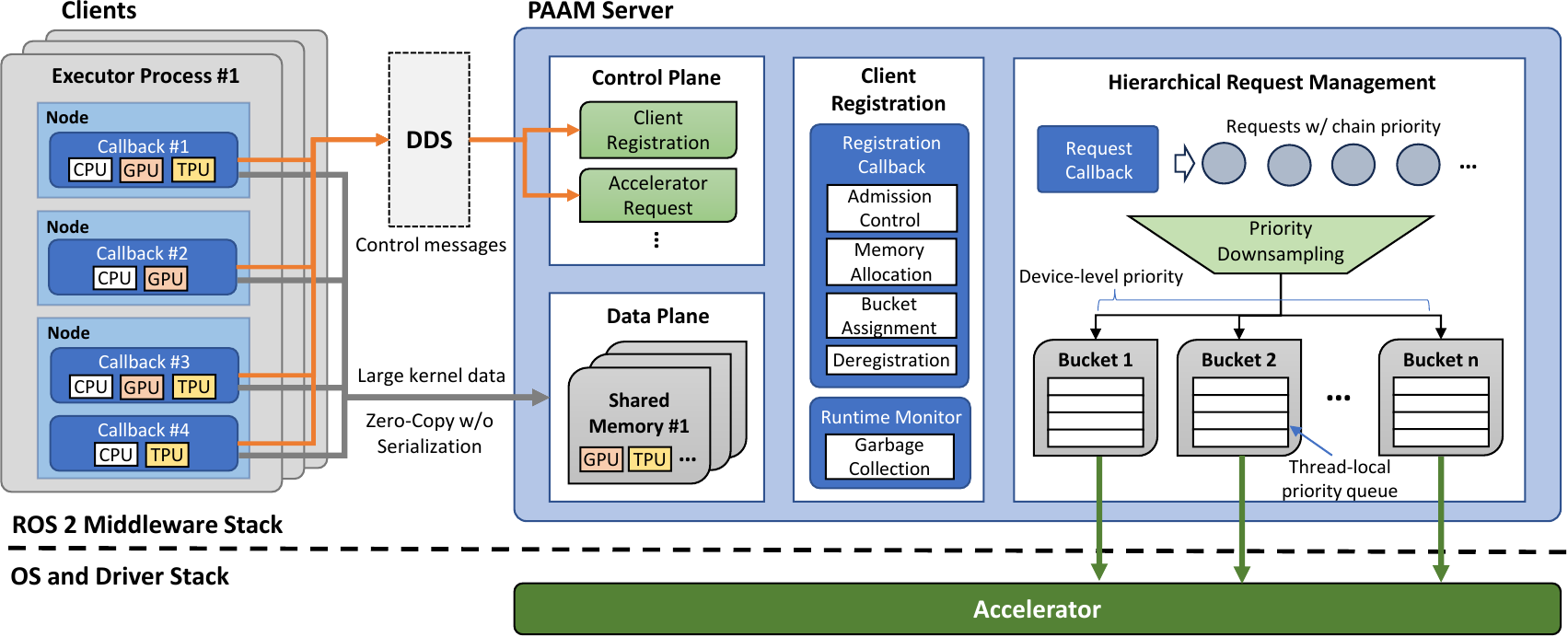}
}
\caption{ROS 2 PAAM Framework}
\label{fig:ROS 2AAMF} 
\end{figure*}

\section{Challenges}\label{sec:challenges}\label{challenges}
Growing adoption of shared accelerators across complex software stacks can compromise real-time guarantees essential for safety-critical operations. This section delves into the challenges that arise from accelerator integration into open and closed-source real-time robotic applications.

\subsection{Priority Inversion and Unbounded Blocking}
For systems that utilize accelerators, requests to use that accelerator from lower priority chains can block those requests from higher priority chains because the OS and device drivers are oblivious to the concept of processing chains and chain priorities in robotic middleware. Moreover, the default driver behavior for many accelerators is to execute requests in FIFO order without regard to chain or executor priorities. In the worst case, under the direct invocation model, where callbacks within a ROS system directly invoke the shared accelerator using the driver API, the accelerator request of the highest-priority chain may be delayed by the requests of all lower-priority chains, causing excessive and violently variable blocking time. None of the existing real-time chain scheduling and analysis for ROS~2~\cite{casini2019response,tang2020response,blass2021ros,blass2021automatic,choi2021picas,sobhani2023} has considered this problem.

\subsection{Poor Accelerator Resource Utilization}
The direct invocation approach causes kernels to be launched from different process contexts. In the case of GPUs, the default device driver behavior interleaves the execution of those kernels from various process contexts but does not execute them concurrently on the hardware~\cite{capodieci2018deadline,bakita2023hardware,yang2018avoiding}. This creates unnecessary GPU context-switching events and usually results in a larger overall execution time for all interleaved kernels. 
While the interleaved execution provided by the device driver prevents starvation from happening, it can be a significant cause of unpredictable timing behavior. Furthermore, if the system includes non-real-time, best-effort kernels that do not fully use the accelerator's resources, the total utilization can be much less than what it could be. In the case of other accelerators, such as the Coral Edge TPU used in this work, the problem could be much more significant since some of them even limit access by multiple processes~\cite{web:edgetpu_limitation}.

\subsection{Disparity in Chain and Executor Priorities}
To address the aforementioned two problems, several solutions have been developed to schedule accelerator requests based on the priority of calling processes, particularly in the context of shared GPUs~\cite{kim2017server,Kim_JSA18,elliott2013gpusync,patel2018analytical}. 
One may think that such solutions, when applied to ROS-based systems, could offer comparable performance enhancements as witnessed in other systems.
However, in the ROS 2 ecosystem, the priority of chains and callbacks does not always align with the priority of the executor process. Such disparity in process and chain priorities leads to conditions where critical callback chains necessitating accelerator resources cannot access them in a timely manner, facilitating unpredictable timing behavior.

\smallskip\noindent\textbf{Goals:} The primary objectives of our work are to minimize the end-to-end response time of a critical chain in the presence of accelerators and to build a framework that makes it amenable to derive a worst-case bound while achieving efficiency.


\section{PAAM Architecture}\label{sec:arch}
PAAM introduces an accelerator management server running as a standalone executor at the ROS 2 application layer and providing accelerator access as a service to clients.\footnote{A client can be seen as a node in program code, but the entity that issues requests at runtime is an executor process.}
Fig.~\ref{fig:ROS 2AAMF} presents an overview of our framework. PAAM creates one server executor per accelerator, each with one or more worker threads. 
In the following, we will present the details of the key components of PAAM: (i) client registration, (ii) data transport with control and data planes, (iii) hierarchical request management, (iv) client/server execution flow control, (v) GPU- and TPU-specific considerations, and (vi) admission control with analytical bounds. 



 

\subsection{Client Registration} \label{sec:client_reg}
\begin{figure}[t]
	\centering
 \vspace{-1mm}
	\subfloat{
		\includegraphics[width=\linewidth]{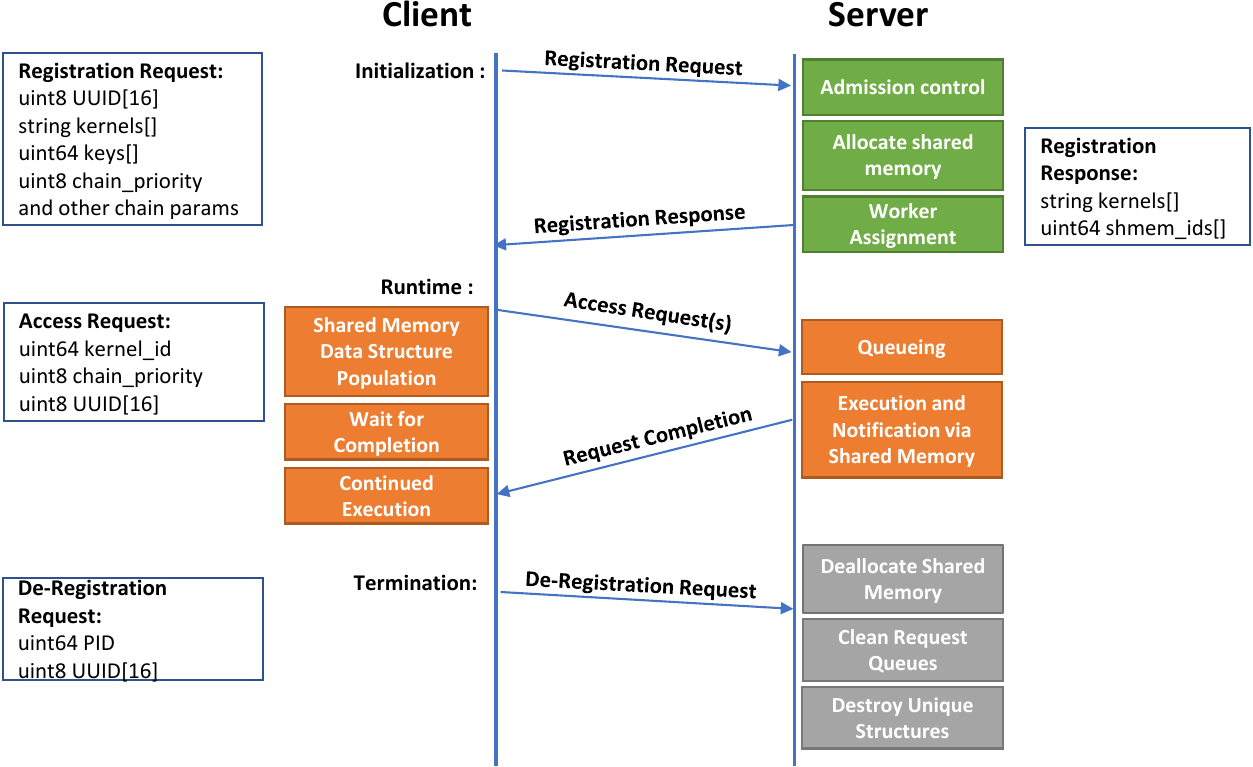}
        }
		\vspace{-1mm}
        \caption{Client Registration Sequence}
	\label{fig:execution-flow}
\end{figure}

PAAM requires that ROS 2 clients that must use accelerator resources register with the server at their startup. This procedure is facilitated through a message exchange sequence that relays the client information: chain priority, accelerator services (kernels) to use,
callback's UUID, and client's executor process ID. An example of a client registration sequence and associated message types is shown in Fig.~\ref{fig:execution-flow}. 
Once a chain registration request is received, the PAAM server performs an admission control test based on the method given in Sec.~\ref{sec:analysis}. If the clients are local to the PAAM server, meaning that they live within the same system, shared memory regions are created and associated with specific accelerator services. 
If remote, only their chain priorities, system UUID, callback UUID, executor PID, and accelerator services are logged. 

\smallskip\noindent\textbf{Mapping Clients to Buckets.} During client registration, the server also maps the chain priority to ``buckets'' that correspond to accelerator-supported priority job queues. We assume that chain priorities are fixed, and the subsequent requests from that callback will be assigned to the same bucket. If this is not the case, however, the clients can simply de-register from the server and re-register with a different chain priority. We discuss more details of buckets and queueing in Sec.~\ref{queueing}. After client registration, the PAAM server responds with shared memory IDs that the client can attach to in order to take advantage of the zero-copy unserialized data plane provided by the PAAM server (Sec.~\ref{sec:data_trans}).

\begin{figure}[t]
	\centering
	\subfloat{
		\includegraphics[width=.9\linewidth]{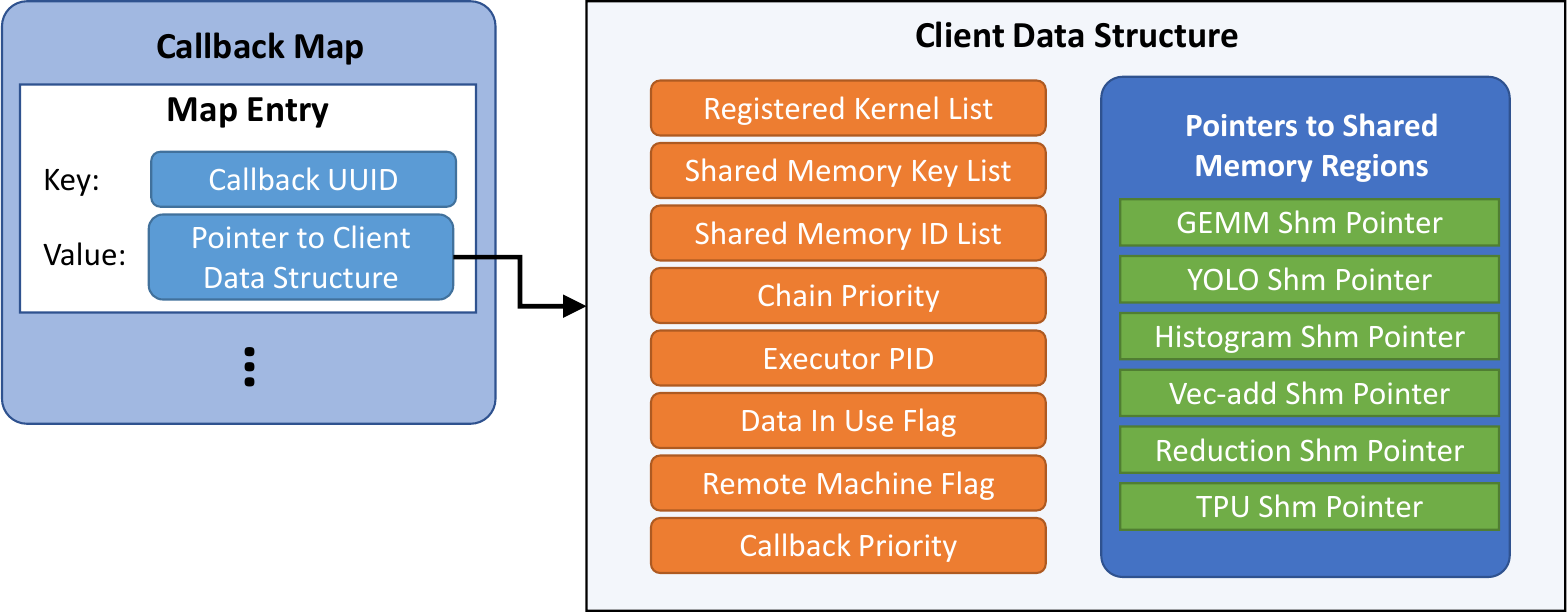}
        }
        \vspace{-1mm}
        \caption{ Server Data Structure of Client Information}
	\label{fig:server_struct}
\end{figure}

\smallskip\noindent\textbf{Client Data Management.}
It is possible for different callbacks of the same client to make requests for the same kernel. To distinguish such requests and prevent data conflicts, PAAM uses callback UUIDs and allocates separate shared memory regions for each of them. In Fig.~\ref{fig:server_struct} we show a diagram of the server-local data structure that is used to keep track of client callbacks within a chain, shared memory regions assigned to them, their anticipated kernels, and other important client metadata.
During the client registration procedure, the server will allocate one client data structure per callback 
requesting accelerators. The structure is then populated, as explained above, with information about the client, as well as pointers to shared memory regions. The server uses an unordered map to link callback UUIDs to client registration structures for use when executing accelerator workloads on specific data. The client registration structure stores a list of applicable pointers to server-generated shared memory regions, sorted by request type, in addition to a list of anticipated kernels, shared memory keys and IDs (used for client de-registration and garbage collection), chain priority, executor process information, data status flags (used to prevent deallocation and destructive kernel cancellation), and a remote machine flag (used to indicate whether the client callback belongs to a networked system).


\smallskip\noindent\textbf{Runtime Monitor.}
With the framework design being targeted at highly dynamic software systems with multiple clients registering, de-registering, and unexpectedly exiting, we have included a runtime monitor for all clients that have registered with the server. When the server is idle, the runtime monitor checks on the status of processes that have registered and will garbage-collect allocated memory that was left stranded by improper nodal de-registration. 
Note that the runtime monitor is independent of the critical path of accelerator workload execution by the PAAM server, and thus runs with non-real-time priority to minimize interference.

\subsection{Data Transport} \label{sec:data_trans}
By default, the ROS 2 environment does not distinguish data and control messages. Such a combined model can lead to excessively large messages being transported via DDS, introducing long transmission time, data serialization, multiple copy operations, and the additional overhead from the ROS-DDS interfacing layer. To mitigate the communication cost between clients and the PAAM server, we opted to establish separate {\em data and control planes} in our framework design. 
This separation ensures that the server can access the necessary data through shared memory without serialization, while control messages are formed in a fixed size and sent over DDS. Furthermore, this approach allows us to bypass the ROS-DDS interface layer, which appears to be a major source of overhead as highlighted in Sec.~\ref{sec:ovh_brkdn},  for large data transmission. 

\smallskip\noindent\textbf{Control Plane.} The control plane involves local client accelerator access request messages of small fixed size (72 bytes) that can automatically leverage zero-copy capabilities of existing DDS implementations such as Iceoryx-enabled~\cite{web:iceoryx} Cyclone DDS~\cite{cyclonedds}. Other variable-sized control-plane messages include client registration requests, server registration responses, and remote client accelerator access requests. 

\smallskip\noindent\textbf{Data Plane.}
In the data plane, we pre-allocate memory based on the maximum data size for each request type during client registration. This approach allows us to manage variable-sized data from callbacks of the same node in shared memory, eliminating the overhead associated with resizing these regions. For local clients, once the server allocates memory and relays the shared memory IDs back to the client, client callbacks can then dynamically attach to and populate input data structures before submitting a request. It is worth noting that, since each shared memory region is unique to each client callback, there is no data conflict even if many callbacks of the same client make the same request type simultaneously. Leveraging shared memory in a separate data transport plane eliminates the copy operations that would otherwise take place in DDS message transport; hence, data can be directly assigned to the pre-established memory regions, without needing to allocate objects or serialize them for the DDS transport. 
While callbacks have to initially write data to shared memory, such a write is necessary for any data-producing callback regardless of PAAM's presence. Upon the completion of a kernel, the server-side kernel wrapper updates a flag in the corresponding shared memory region, notifying the client callback of the kernel completion and enabling it to resume its execution path. 

\begin{figure}[t]
	\centering
	\subfloat{
		\includegraphics[width=0.45\linewidth]{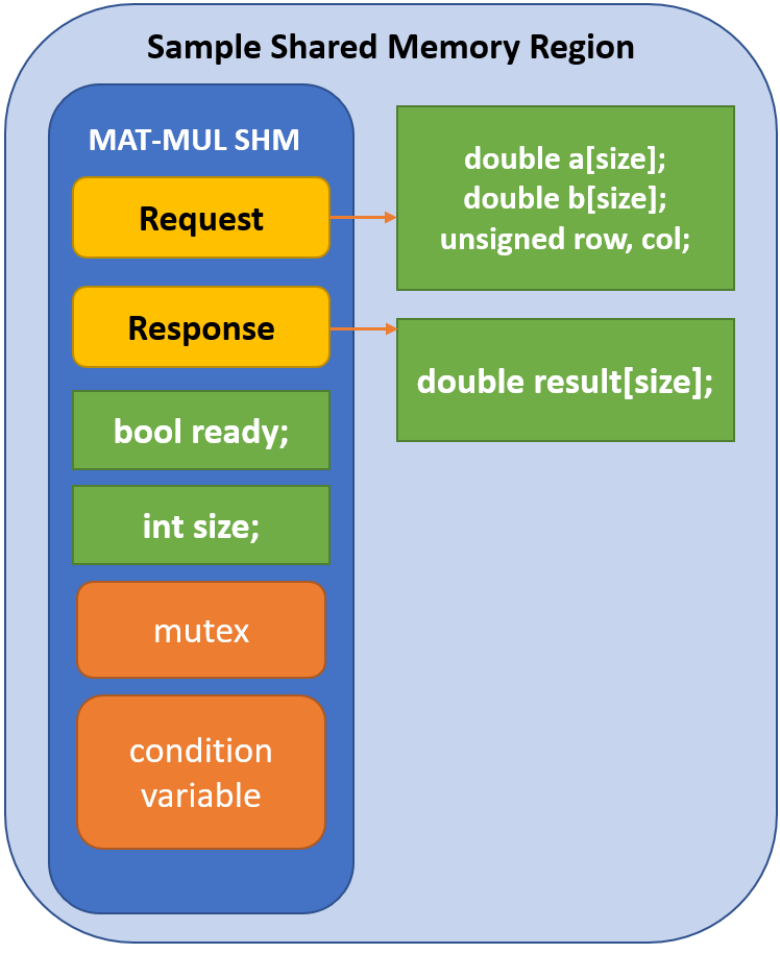}
        }
	\vspace*{-2mm}
        \caption{ Sample Shared Memory Region}
	\label{fig:sample_shm}
\end{figure}

In Fig.~\ref{fig:sample_shm} we describe a sample shared memory region that is allocated by the server and attached to a client. The shared memory region maintains distinguished request and response data regions, allowing for a simple programming interface that is consistent across the client and server programs. Each shared memory region also contains locks for data protection as well as condition variables used to awaken sleeping client callback threads. The ready flag is used to verify that the result from the PAAM server is ready, especially if the client experiences a spurious wakeup. It is also worth noting that these data structures are customizable and can be tailored to any type of local request using the PAAM framework.

\smallskip\noindent\textbf{Remote Clients.}
The use of shared memory in the data plane is infeasible in this case. Instead, we rely on standard serialization and DDS message transport of variable-sized data structures over network sockets. While this incurs higher overhead than the shared memory-based data plane, it allows for accelerator-less devices to remotely utilize accelerator resources for heavy processing workloads, with their chain priorities still being respected.
  

\subsection{Two-level Hierarchical Request Management}\label{queueing}
The PAAM architecture employs a two-level hierarchical prioritization structure for incoming request (accelerator job) queue management: buckets and thread-local queues. To leverage hardware-supported prioritization and preemption, the PAAM server creates $n$ worker threads, which we call {\em buckets}, where $n$ is the number of priority levels provided by the associated accelerator hardware, e.g., CUDA on Nvidia Jetson AGX Xavier provides up to six stream priorities. While not many accelerators except for GPUs support device-level prioritization and preemption as of yet, the computer architecture community is recently introducing such capabilities in NPUs~\cite{choi2020prema}. The number of buckets is specific to accelerator hardware and 
is gathered during the server initialization phase. Within each bucket, there is a thread-local priority queue to determine execution order in accordance with the corresponding chain's priority. Each bucket has access to the server executor process memory, including all of the shared memory regions assigned during the registration process of clients. With this structure, we can efficiently reap the benefits of our zero-copy data plane for local clients when executing accelerator services on client-assigned data.\par 
Upon receiving an accelerator access request from a client, the PAAM server will determine the appropriate bucket to handle the incoming request. Original bucket assignment is performed during the client registration process and is determined through a simple down-sampling heuristic that divides chain priorities into $n$ evenly sized groups; the highest priority chains are assigned the highest priority buckets. 
If $n=1$, i.e., the associated accelerator does not provide any device-level priority and preemption, all requests are handled by the thread-local priority queue of a single worker thread. 
If the multi-bucket (worker) implementation for multi-stream capable accelerators is utilized, the requests that are run on a higher priority stream can preempt those executing on lower priority streams. If all requests are assigned to the same bucket, then preemption cannot happen. This will be the case for other accelerators that do not support multiple hardware priority levels, such as TPUs. Our analysis for admission control in Sec.~\ref{sec:analysis} considers such preemption among buckets and blocking within the same bucket. During runtime, the chain priority is cached on the server and can be quickly referenced to rapidly transport the request into the right bucket's job queue. The execution and subsequent completion notification of the accelerator service is explained in more detail in Sec.~\ref{flows} \par

\subsection{Client and Server Execution Control Flows}\label{flows}
\begin{figure}[t]
	\centering
	\subfloat[Synchronous Request]{
		\includegraphics[width=.7\columnwidth,align=c]{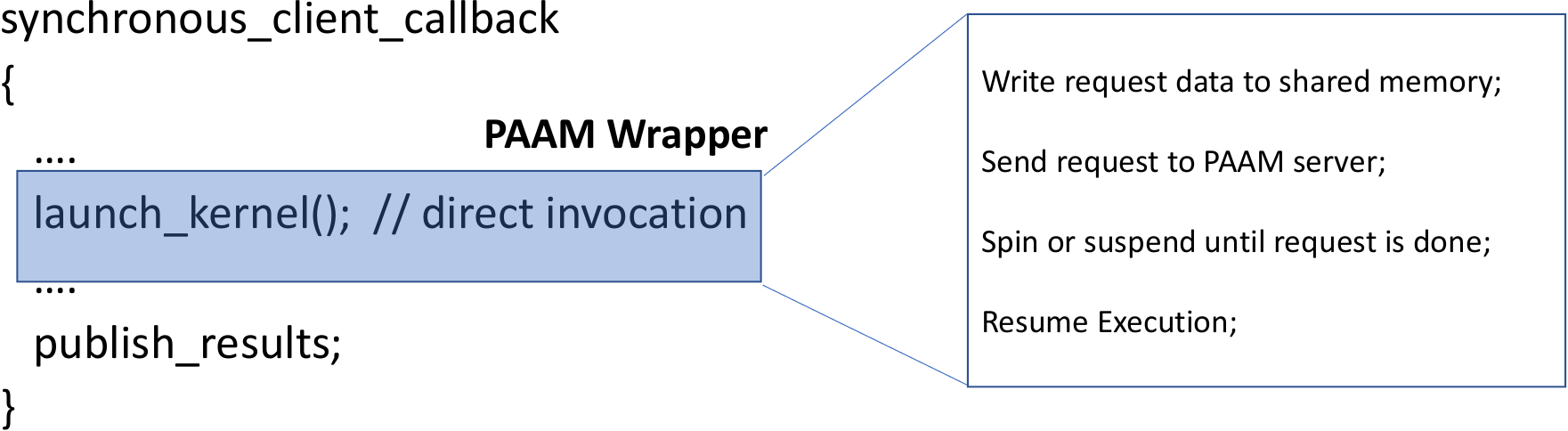}\label{fig:sync_wrapper}
	}\\\vspace{-2mm}
 	\subfloat[Asynchronous Request]{
		\includegraphics[width=.7\columnwidth,align=c]{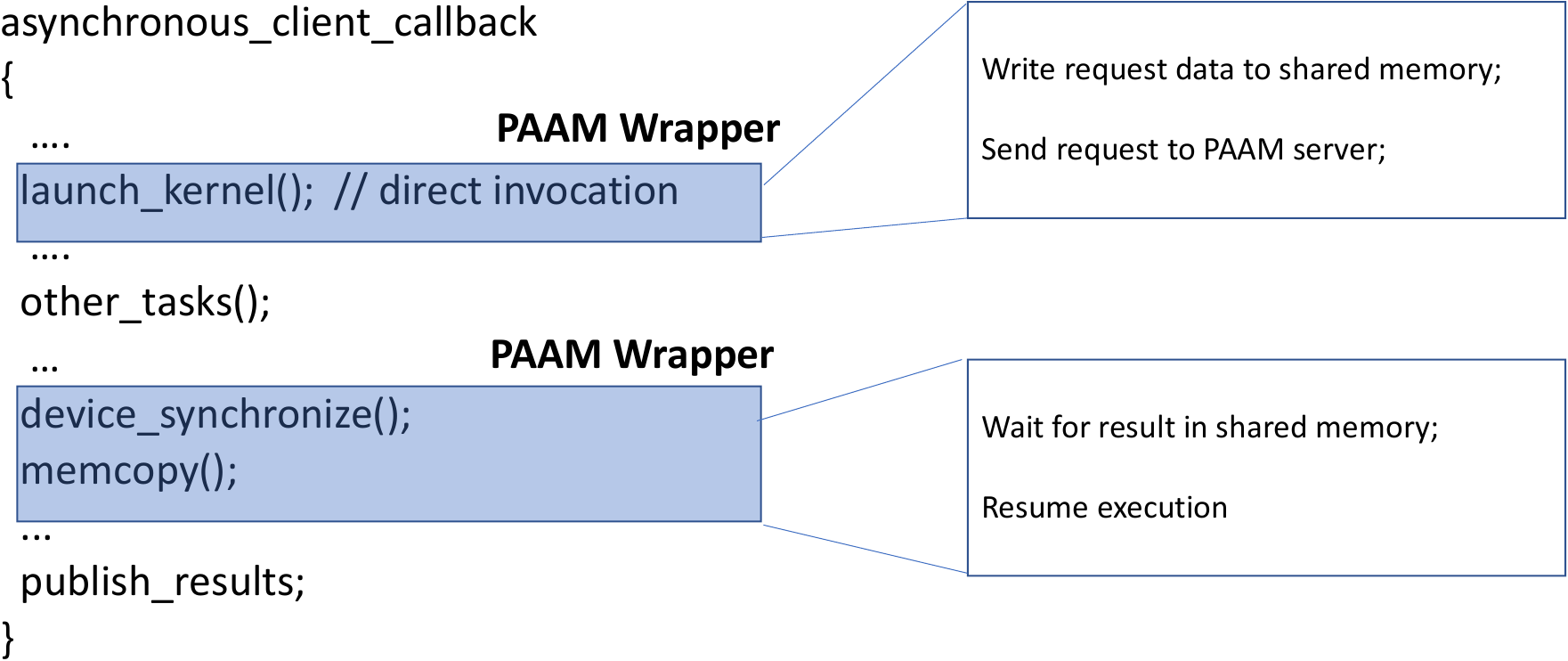}\label{fig:async_wrapper}
	}
	\vspace{-1mm}
	\caption{PAAM Wrapper for Callback Accelerator Requests}
	\label{fig:aamf_wrappers}
\end{figure}

PAAM preserves the client's original execution control flows to minimize code changes. Callbacks can directly make a request as if it were making a direct invocation of an accelerator resource. In Fig.~\ref{fig:sync_wrapper} and \ref{fig:async_wrapper}, we show that the execution flow for synchronous and asynchronous accelerator invocation performed by client callbacks is not disrupted by the use of PAAM. Our wrappers assign data to shared memory regions, encapsulate the control plane messages, and send the requests to the PAAM server. For synchronous requests, the client executor either suspends or spins on a status flag in shared memory that serves as a notification from the server. Upon receiving a notification from the server, the client will resume its normal execution. For asynchronous callbacks, the client can perform other CPU-based workloads that are not dependent on the accelerator result while the PAAM server is handling the request.

 \begin{figure}[t]
	\centering
 \vspace{-2mm}
	\subfloat{
		\includegraphics[width=0.65\linewidth]{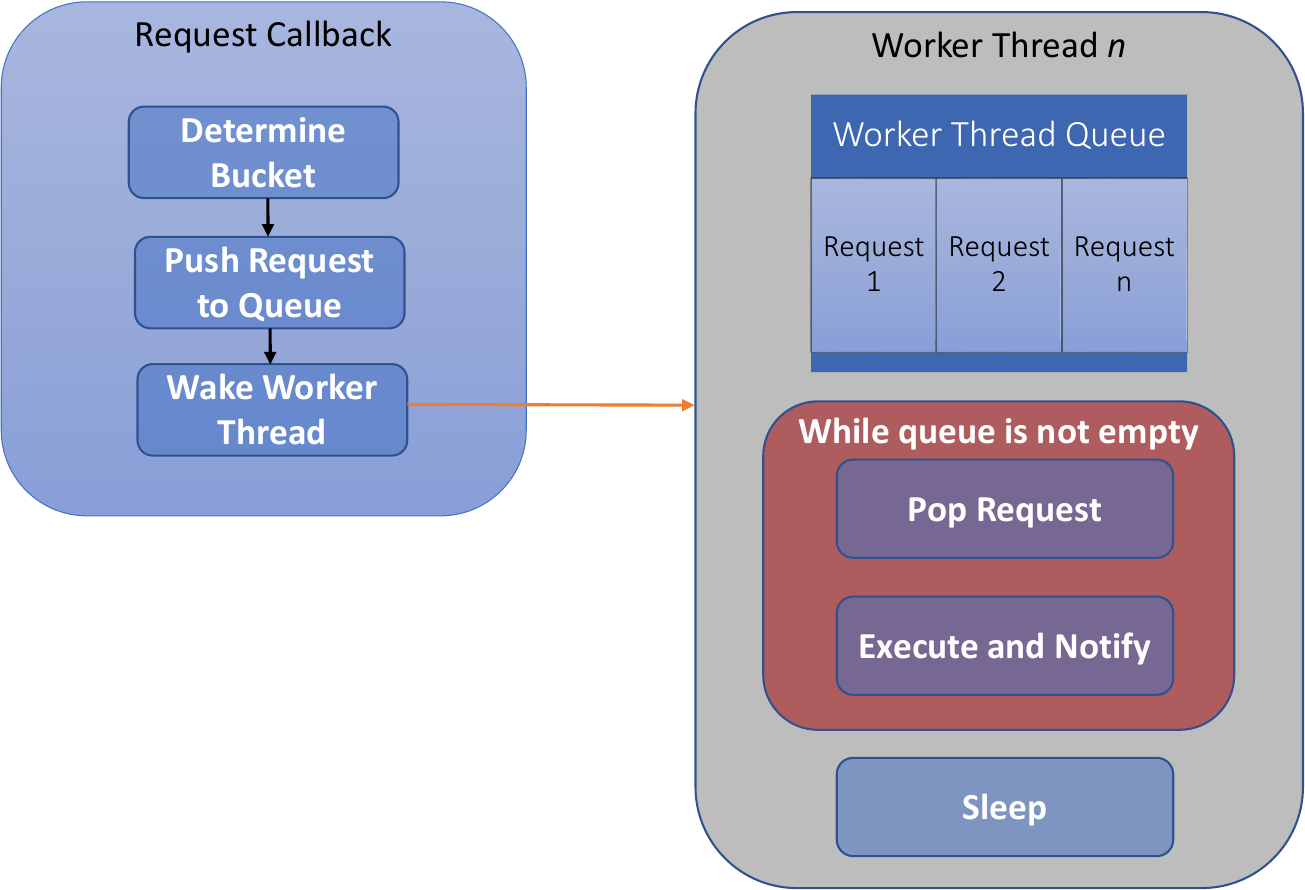}
	}
	 \vspace{-2mm}
	\caption{Server Execution Flow}
	\label{fig:server_flow}
\end{figure}

In the server process, as shown in Fig.~\ref{fig:server_flow}, we show a simple structure of how a client request is enqueued into the proper bucket queue and executed in chain priority order. 
Upon receiving a request from a client callback, the incoming request callback on the server will immediately determine the worker thread that the request should belong to, enqueue the request to that worker's local request queue, and wake the worker thread if it was previously sleeping. The worker thread then pops the request from its thread-local queue and executes it. 
Once the execution of the accelerator service is complete, the server either wakes up the client executor or updates the status variable in shared memory -- indicating that the result in shared memory is ready. When waking up the clients, the server will leverage the mutex and condition variable associated with the client callback that was allocated during client registration and stored in shared memory as shown in Fig~\ref{fig:sample_shm}. 
\subsection{GPU- and TPU-specific Considerations} \label{gpu_consid}
\subsubsection{GPU}
For accelerators that support hardware-level prioritization of kernels (e.g. GPUs), we establish buckets (worker threads) for {\em prioritized streams} that can preemptively execute kernels and reduce blocking time, without detracting from the dependability of the application. 
In the case of Nvidia GPUs, each bucket corresponds to a unique CUDA stream with a unique stream priority. Recall that, during client registration, clients are assigned to buckets based on their callback chain priorities, which effectively down-samples the chain priorities into CUDA stream priorities as discussed in Sec.~\ref{queueing}. If there are two or more kernels assigned to the same bucket with the same chain priority, PAAM executes them one at a time in arrival order. Hence, blocking time still exists, but this approach of associating prioritized streams to buckets allows us to exploit the preemptive mechanisms provided by driver APIs. In this way, the kernels executed by buckets associated with higher chain priorities can preempt any executing kernels from lower-priority buckets. 

For balanced real-time performance and resource efficiency, we allow the concurrent execution of kernels~\cite{yang2018avoiding,otterness2017evaluation,xiang2019pipelined,amert2017gpu} in the lowest priority bucket such that non-critical or best-effort chains can better utilize GPU's internal compute resources at the expense of possible non-deterministic slowdowns in their execution. This approach, however, does not cause slowdowns to kernels in higher-priority buckets because of stream-level preemption. By leveraging a single accelerator context maintained by the PAAM design, we utilize device driver features that interleave instructions for multiple processes' requests without requiring an expensive change of the GPU context or using an additional closed-source layer like Nvidia MPS~\cite{nvidia_mps}.\footnote{As of this writing, MPS is still not available for the ARM architecture.}

\subsubsection{TPU}
To showcase PAAM's compatibility with multiple types of accelerators, we also integrated the Coral Edge TPU with PAAM. Unlike Nvidia GPUs, the Coral TPU provides no hardware-level concurrency prioritization. In addition, it can only support one process context and invocation at any given time~\cite{web:edgetpu_limitation}, meaning that other processes cannot use the TPU until the one that is currently using the TPU terminates. This could be a significant limitation in multi-process robotic environments, effectively limiting the utilization of the TPU to one client executor. PAAM solves this issue and allows multiple clients to seamlessly share the TPU at runtime. 

The TPU-specific design considerations include a single worker (bucket) for handling the requests for TPU usage. This simplifies the design of the worker to include a non-preemptive and non-concurrent priority-based execution order. Similar to the design of the GPU workers, we create a single worker with a priority queue for incoming requests that is pinned to a separate CPU separate from the GPU workers and maintains real-time priority in Linux.

\subsubsection{Managing Multiple Accelerators}
If the system has multiple units of the same accelerator type, e.g., 2 TPUs, the PAAM server creates sets of workers, 
where each set corresponds to one accelerator unit. 
The assignment of each callback/accelerator segment to a specific unit of the requested accelerator type is performed at the registration phase using bin-packing heuristics such as WFD. 
This approach is analogous to partitioned multiprocessor scheduling,
and help improve the responsiveness of critical chains by assigning them to a separate, designated accelerator. 

\subsection{Admission Control}\label{sec:analysis}
For any new client, the PAAM server performs admission control to determine its acceptance. One approach would be to simply assess the utilization of clients' accelerator requests. 
However, to guarantee end-to-end performance,
we derive the {\em worst-case response time of a chain} that includes callbacks accessing accelerators through PAAM. As mentioned in Sec.~\ref{sec:background}, we adopt the priority-driven chain execution mechanism of PiCAS~\cite{choi2021picas}. Hence, our approach extends the analysis from \cite{choi2021picas}, incorporating the delays induced by accelerator access. While PAAM supports various client executors, for analysis purposes, we follow the same assumptions as in \cite{choi2021picas}: clients use single-threaded executors, each statically assigned to a CPU core with a distinct real-time process priority. Our analysis focuses on chains with constrained deadlines.

Let us consider a chain $\Gamma_c$ executing in one executor. This chain could be a sub-chain of a larger chain $\Gamma^*$ that spans across multiple executors on different CPU cores. We will first upper-bound the response time of $\Gamma_c$ and then show how the end-to-end response time of $\Gamma^*$ can be obtained. 

In PAAM, the key behavior rules that affect the request handling time for accelerator access are summarized as follows.
\begin{itemize}[leftmargin=*]
	\item R1: For each accelerator device, one server is created and runs on a dedicated CPU. Hence, different PAAM servers do not interfere with each other on both CPU and accelerator.
	\item R2: When an PAAM server receives a request, it is sent to one of $n$ buckets through priority downsampling and handled by the corresponding worker thread of the server. 
	\item R3: The worker thread executes requests in its bucket non-preemptively in the order of requests' chain priority.
	\item R4: If the server has multiple buckets, requests in a higher-priority (HP) bucket can preempt those in a lower-priority (LP) one.
\end{itemize}
R1 allows us to analyze the request handling time for one accelerator independent of other accelerators in the system. R2-R4 are resulted from our two-level hierarchical request management. By R2 and R3, a chain $\Gamma_c$'s request can be delayed by up to one request from an LP chain that the accelerator is already executing. R4 adds interference from HP chains during the handling time of $\Gamma_c$'s request.
In addition, there are two types of overhead to consider: (i) $\epsilon$, which is the overhead imposed on every request (accelerator segment) by the PAAM server ($\delta_c$ segments for $\Gamma_c$), and (ii) $\kappa$, which is the device-level preemption cost that can occur before and after each segment (at most $2\kappa$ per segment). We define $A_{i,j}^*=A_{i,j}+2\kappa$ to incorporate the preemption delay into the execution time of the $j$-th accelerator segment of a callback $\tau_i$. Based on this, the maximum time to complete all accelerator requests of the chain $\Gamma_c$ with overhead is given by:
\begin{equation}
    \label{eq:beta}
    H_{c}^* = H_c + 
     \delta_c\cdot \epsilon 
\end{equation}
where $H_c$ is the cumulative request handling time for all accelerator segments of $\Gamma_c$ under PAAM (incl. delay R2-R4).

Before bounding the handling time $H_c$, let us analyze how many accelerator requests an PAAM server can receive from an interfering segment $\tau_{k,q}$ during an arbitrary time interval $t$. 

\begin{lemma}[arrival bound]\label{lm:req_arrival} 
	The maximum number of requests that an accelerator segment $\tau_{k,q}$ from a schedulable chain $\Gamma_{c'}$ can generate in an arbitrary interval $t$ is bounded by
	\begin{equation}\label{eq:req_arrival}
		\begin{split}
			\mu_{k,q}(t) & \gets
			\Big\lceil \frac{t}{T_{c'}} \Big\rceil +1 
		\end{split}
	\end{equation}
\end{lemma}
\begin{proof}
The requests of a segment $\tau_{k,q}$ are made in a periodic manner based on its chain period, with a possible jitter $j_1<T_{c'}$ due to that $\tau_{k,q}$ can start as soon as its preceding segment or callback completes. If the interval $t$ is aligned with the release time of the chain $\Gamma_{c'}$, this gives $ \lceil \frac{t+j_1}{T_{c'}}\rceil$ (recall $\mu()$ is not a response time test; it is to bound the number of arrivals of requests during $t$). However, as $t$ is an arbitrary interval, there can be {\em carry-in} execution of $\tau_{k,q}$ in the interval $t$ due to delayed execution. For instance, the start of the execution of $\tau_{k,q}$'s job can be delayed by other callbacks on the same executor (e.g., blocking or self-pushing effect of non-preemptive scheduling) or by OS-level preemption from HP executor processes on the same CPU core. While this may introduce a more number of jobs of $\tau_{k,q}$ in the interval $t$ than $\lceil \frac{t}{T_{c'}}\rceil$, it is known to be limited to most once in a schedulable system with the constrained deadline model~\cite{bertogna2008schedulability} and its effect can be taken into account as a jitter, i.e., $j_2<T_{c'}$. The sum of these jitters $j_1$ and $j_2$ cannot exceed $T_{c'}$ since $\Gamma_{c'}$ is schedulable, and therefore $ \lceil \frac{t+j_1+j_2}{T_{c'}}\rceil\le \lceil \frac{t}{T_{c'}}\rceil+1$. It is worth noting that the well-known self-pushing effect of non-preemptive scheduling can be upper bounded by at most one carry-in job in the interval $t$ because $\Gamma_c'$ is schedulable. 
\end{proof}

To bound $H_c$, we follow the double-bounding approach~\cite{Kim_JSA18,kim2021addressing} that derives two separate but safe bounds and takes the minimum between them. 
We will present each approach in a respective subsection and then derive the response time analysis of a chain. We will use the following notation:
\begin{itemize}
	\item $hps(\tau_{i,j})$ (and $lps(\tau_{i,j})$): the set of accelerator segments that are from HP (and LP) chains than the chain of $\tau_{i,j}$ and use the same accelerator as $\tau_{i,j}$. 
	\item $b(\tau_{i,j})$: the bucket of the PAAM server assigned to $\tau_{i,j}$. 
\end{itemize}

\subsubsection{{Bound by Per-segment Handling Time}}
First, the handling time $H_c$ of a chain can be obtained by calculating the handling time of each accelerator segment $\tau_{i,j}$, $H_{c,i,j}$, and then adding them up, i.e., $H_c=\sum_{\tau_i\in\Gamma_c}\sum_{1\le j \le \eta_i}H_{c,i,j}$. 
\begin{lemma}[handling time per segment]  
The maximum handling time of the $j$-th accelerator segment of a callback $\tau_i \in \Gamma_c$ under PAAM is upper-bounded by the recurrence:
\begin{equation}\label{eq:1}
\resizebox{.94\linewidth}{!}{$
\begin{split}
    H_{c, i, j} & \gets A_{i,j}^* + \!\!\!\!\!\!\!    
    \max_{\substack{\tau_{k,q} \in lps(\tau_{i,j}) \\\wedge b(\tau_{k,q}) = b(\tau_{i,j})}}{\!\!\!\!\!\!A_{k,q}^*}     
    +\!\!\!\!\!\sum_{\substack{\tau_{k,q} \in hps(\tau_{i,j})}}{\!\!\!\!\!\!\!\!\!\! \mu_{k,q}(H_{c,i,j}) \! \cdot \! A_{k,q}^*}
\end{split}$}\hspace{-10pt}
\end{equation}
The recurrence starts with the first two terms. 
\label{lm:max_waiting}
\end{lemma}
\begin{proof}
As explained earlier, the handling time $H_{c,i,j}$ of the accelerator segment of $\tau_{i,j}$ is determined by three factors. First, $\tau_{i,j}$'s own execution time inflated with the preemption cost, $A_{i,j}^*$. Secondly, blocking time occurs when the accelerator is already handling a request from an LP chain in the same bucket, which is the first term of the equation. Note that those in other buckets do not cause blocking because they can be preempted (recall PAAM creates more than one bucket only when the accelerator supports device-level preemption). Thirdly, an additional delay is imposed by HP requests during the handling time of $H_{c,i,j}$, the number of which is bounded by $\mu_{k,q}(H_{c,i,j})$. Therefore, the delay caused by requests from $\Gamma_h$ is bounded by { $\sum\mu_{k,q}(H_{c,i,j})\cdot A_{k,q}^*$}, and the maximum handling time can be obtained by adding these three factors.
\end{proof}

\subsubsection{{ Bound by Per-chain Handling Time}}
Unlike the first approach, the second approach takes into account the total amount of interfering requests during the chain $\Gamma_c$'s execution. 

\begin{lemma}[handling time per chain job]
The maximum handling time of all accelerator segments from all callbacks of a chain $\Gamma_c$ under PAAM is upper-bounded by:
\begin{equation}\label{eq:2}
\resizebox{.95\linewidth}{!}{$
\begin{split}
    H_{c} & \!\gets\!\!\!
    \sum_{\tau_{i,j}\in\Gamma_c}\!\!\!
    \Big(\! A_{i,j}^*+\!\!\!\!\!\! \max_{\substack{\tau_{k,q} \in lps(\tau_{i,j}) \\\wedge b(\tau_{k,q}) = b(\tau_{i,j}) }}{\!\!\!\!\!\!A_{k,q}^*} 
    \Big) 
    \! + \!\!\!\!\!\!\!\!\!\!\!\sum_{\substack{\quad\tau_{k,q}\in \!\!\!\!\!\bigcup\limits_{\tau_{i,j}\in \Gamma_c}\!\!\!\!\!\! hps(\tau_{i,j})}}{\!\!\!\!\!\!\!\!\!\!\!\!\!\!\mu_{k,q}(R_c)\! \cdot \! A_{k,q}^*}
\end{split}$}\hspace{-10pt}
\end{equation}
where $R_c$ is the response time of $\Gamma_c$. This can be solved during the iterative calculation of $\Gamma_c$'s response time $R_c$ given next. 
\label{lm:max_waiting2}
\end{lemma}
\begin{proof}
Each accelerator request made by a segment $\tau_{i,j}$ of $\Gamma_c$ may experience blocking time from an LP request in the same bucket, and the total blocking time for $\Gamma_c$ can be bounded by the summation of each blocking time for all segments of $\Gamma_c$, as shown in the first term that also includes the execution time of $\Gamma_c$'s own requests. During $\Gamma_c$'s response time, the number of requests from an HP chain $\Gamma_h$ is bounded by $\mu_{k,q}(R_c)$, and the second term of the equation bounds the delay from HP chains.
\end{proof}

It is worth noting that the two approaches given in Lemmas~\ref{lm:max_waiting} and \ref{lm:max_waiting2} do not dominate each other. While the blocking time from LP chains is the same, the amount of interference from HP chains is different. For instance, if the number of accelerator segments of a chain $\Gamma_c$ under analysis is relatively small, Lemma~\ref{lm:max_waiting} can give a tighter bound since it only considers HP interference during the segments' handling time; in the opposite case, Lemma~\ref{lm:max_waiting2} can give a better result since its second term is not subject to the number of segments $\Gamma_c$ has. Since both approaches upper bounds the handling time, we can take the minimum between them to obtain the handling time $H_c$, and use this in Eq.~\eqref{eq:1} to compute $H_c^*$, the time to complete all accelerator requests of $\Gamma_c$ with overhead.

As mentioned before, we analyze a chain or sub-chain $\Gamma_c$ that has all of its callbacks running on a single executor, and then 
discuss how chains spanning across multiple executors can be analyzed.
We use the following notation to specify other interfering chains:
\begin{itemize}
	\item $hp(\Gamma_c)$ (and $lp(\Gamma_c)$): the set of HP (and LP) chains than $\Gamma_c$ running in the same executor as $\Gamma_c$. 
	\item $hpp(\Gamma_c)$: the set of chains running in executors on the same CPU as $e(\Gamma_c$) and with higher process priority than $e(\Gamma_c)$.
\end{itemize}
Let us first review the response time test of PiCAS~\cite{choi2021picas}, the priority-driven execution mechanism adopted in PAAM. 
\begin{lemma}[chain with no accelerator segment~\cite{choi2021picas}]\label{lm:wcrt_picas}
The worst-case response time (WCRT) of a chain $\Gamma_c$ without any accelerator segment under the priority-driven chain execution mechanism is bounded by the following recurrence: 
\begin{equation}\label{eq:wcrt_picas}
\resizebox{.94\linewidth}{!}{$
\begin{split}
    R_{c} \gets & \mathcal{B}_c  +\mathcal{E}_c
    + \!\!\!\! \sum_{\substack{\Gamma_h \in hp(\Gamma_c)}} \!\!\!\!\!\! \mu_h(R_c) \cdot \mathcal{E}_h 
    + \!\!\!\! \sum_{\Gamma_h \in hpp(\Gamma_c)}\!\!\!\!\!\! \mu_h(R_c)\cdot  \mathcal{E}_h 
\end{split}$}\hspace{-10pt}
\end{equation}
where $\mathcal{B}_c=\max_{\Gamma_l\in lp(\Gamma_c)}\max_{\substack{\tau_j \in \Gamma_l}} E_j$ is the blocking time from an LP chain, 
$\mathcal{E}_c =\sum_{\substack{\tau_i\in \Gamma_c }}{E_i}$ is the sum of WCET of all callbacks of $\Gamma_c$, and $\mu_h(R_c)=\lceil \frac{R_c}{T_h} \rceil +1 $ is analogous to Lemma~\ref{lm:req_arrival}. The recurrence starts with the first two terms.
\end{lemma}
In the above analysis, the first term captures blocking time due to the non-preemptive callback scheduling of ROS 2. As the priority-driven execution mechanism prevents further execution of callbacks from LP chains until $\Gamma_c$ completes, there can be at most one blocking from a callback of an LP chain on the same executor. The second term $\mathcal{E}_c$ is the execution time of $\Gamma_c$ itself. The third term is the interference from HP chains on the same executor. An interesting part is the fourth term. This captures the fact that the executor process of $\Gamma_c$ can be preempted by other HP executor processes running on the same CPU core (i.e., OS-level preemption), each of which is activated by the arrival of any chain it has. 

We now derive the response time of a chain $\Gamma_c$ with accelerator segments under the PAAM framework by extending the above analysis. The additional factors we need to consider are: (i) accelerator segment execution time $H_c$ of the chain $\Gamma_c$ under analysis, (ii) interference due to accelerator usage of HP chains on the same executor, and (iii) executor-level interference from other executors on the same CPU core.

\begin{theorem}[PAAM]
The worst-case response time (WCRT) of a chain $\Gamma_c$ with accelerator segments under the PAAM framework is bounded by the following recurrence: 
\begin{equation}
\begin{split}
R_{c} &\gets  \mathcal{B}_c+\mathcal{E}_c+H_c^*
    + \sum_{\substack{\Gamma_h \in hp(\Gamma_c)}}\mu_h(R_c)\cdot (\mathcal{E}_h+H_h^*) \\ 
    &+\sum_{\Gamma_h \in hpp(\Gamma_c)}\mu_h(R_c)\cdot ( \mathcal{E}_h +spin(\Gamma_h))
\label{eq:wcrt_chain}
\end{split}
\end{equation}
where $spin(\Gamma_h)= H_h^*$ if $\Gamma_h$ spins (busy waits) on the CPU while waiting for the completion of accelerator requests, and $spin(\Gamma_h)=
\delta_h\cdot \epsilon$ otherwise ($\Gamma_h$ suspends). The recurrence starts with the first three terms, i.e., $R_{c} = \mathcal{B}_c + \mathcal{E}_c + H_c^*$.
\label{lm:wcrt_task}
\end{theorem}
\begin{proof}
The equation is an extension of Eq.~\eqref{eq:wcrt_picas}. The term $H_c^*$ is the handling times for accelerator segments of $\Gamma_c$ itself. For HP chains running in the same executor as $\Gamma_c$ (the first summing term), additional interference caused by each instance of an HP chain $\Gamma_h$ due to its accelerator execution time $H_h^*$ needs to be considered. This is because, while any accelerator request of a callback of $\Gamma_h$ is being handled by PAAM, this callback remains uncompleted and the executor cannot execute any other callback. On the other hand, for chains running on other executors with higher process priority on the same CPU core, the amount of additional interference per chain instance depends on whether that chain spins or suspends during accelerator requests.
In case of spinning, the executor actively consumes CPU cycles, so all components of $H_h^*$ given by Eq.~\eqref{eq:beta} need to be taken into account. In case of suspension, the corresponding executor suspends and the interference is limited to  $\delta_h \cdot \epsilon$, which is the total overhead caused by PAAM. In the literature, self-suspending behavior of an HP task is known to introduce additional penalty to its LP tasks and such penalty can be bounded by a jitter term~\cite{bletsas2015errata,chen2019many}. As $\mu_h(R_c)$ already takes into account the maximum possible jitter for the constrained deadline model, no additional penalty needs to be considered.
As there is no other source of interference or blocking delay possible for $\Gamma_c$, the summation of these terms can upper bound the response time.
\end{proof}
As known in the literature~\cite{von2021realistic,davis2010quantifying}, a sufficient schedulability test for non-preemptive scheduling can be obtained by adding the blocking term to the response time test for preemptive scheduling, and our analysis follows this approach. This is safe but pessimistic. We believe a tighter (or exact if possible) analysis could be derived by considering the level-$i$ active period that can precisely capture the self-pushing phenomenon of non-preemptive scheduling.

\medskip\noindent\textbf{Processing Chain over Multiple Executors.}
Using Theorem~\ref{lm:wcrt_task}, we can obtain the end-to-end latency of a chain $\Gamma^*$ that consists of multiple sub-chains $\Gamma_c\in \Gamma^*$, { each of which is executed by different executors.} This can be directly computed by adding all the response times of sub-chains with {  communication cost $\epsilon'$ ($ \epsilon'< \epsilon$; see Sec.~\ref{sec:eval} for the breakdown of the PAAM overhead $\epsilon$)}, i.e., $R^*=\sum_{\Gamma_c\in \Gamma^*}R_c+\epsilon$. This works because, although subsequent sub-chains may experience release jitters when the preceding sub-chains finish earlier than their WCRT, the negative impact of such jitters on LP chains has been already considered as a carry-in ({ ``+1'' in Lemma~\ref{lm:req_arrival}}). This approach has been used by other prior work for ROS 2~\cite{choi2021picas, casini2019response}.

\section{Evaluation}\label{sec:eval}
In this section, we assess the effectiveness of our proposed framework over the state-of-the-art and explore the performance characteristics and associated overhead. The evaluation was done on the Nvidia Jetson AGX Xavier platform that has 1 integrated GPU and 1 Google Coral Edge TPU. We configured all CPU and accelerator cores to run at the maximum frequency and disabled dynamic clock frequency scaling to minimize measurement fluctuations. Each experiment was run, uninterrupted, for 10 minutes at a time. \footnote{The source code of our implementation is available at \url{https://github.com/rtenlab/reference-system-paam.git}.} 

\begin{figure}[t]
	\centering
	\subfloat{
		\includegraphics[width=\linewidth]{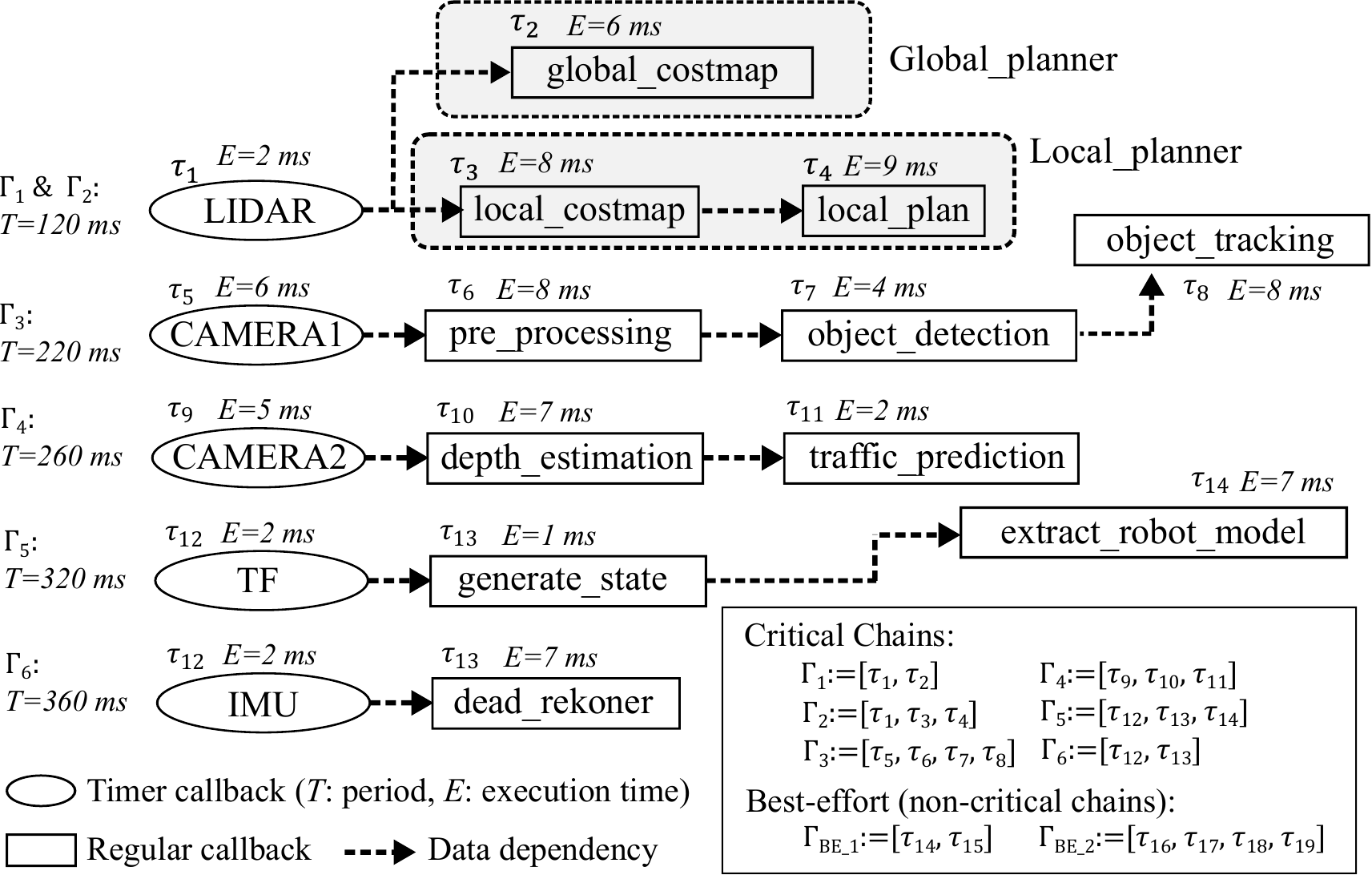}
        }
        \caption{Chain configuration of GPU-enabled robotic system}
	\label{fig:case_study}
\end{figure}

\subsection{Case Study 1: GPU-enabled Robotic System} \label{sec:case_study}
\noindent\textbf{System Setup.}
This scenario is derived from the PiCAS case study~\cite{choi2021picas}, which is inspired by the F1/10 robotic platform, but introduces additional accelerator segments. 
The chain configuration of this scenario can be found in Fig.~\ref{fig:case_study}.
There are 8 processing chains, divided into two categories: critical chains (chains 1-6) and non-critical or best-effort chains (BE~1-2). {For the critical chains, chain 1 has the highest priority, while chain 6 has the lowest priority. } 
Each callback contains one GPU segment with the WCET of 10 ms, i.e., $\forall \tau_i, A_i=10$. 
Callbacks have CPU workloads with various execution times, depicted by $E$ in the figure. 
To compare the maximum end-to-end latency observed from experiments against the worst-case bound from our analysis, we applied the analysis' assumptions to this case study: (i) the PAAM server and its {six} worker threads were pinned to core 0 with real-time priority, and (ii) chain clients were executed by multiple single-threaded executors pinned to other cores. Specifically, the chains were partitioned into four executors, in the same manner as in \cite{choi2021picas}; the executors including critical chains 1-6 were pinned to cores 2-7; the BE chains were assigned to separate executors pinned to cores 2-3 to stress high-criticality chains. BE 1 and 2 are duplicates of chains 1 and 3, respectively. Hence, the BE chains are also periodic, but their response times do not need to be within their periods. 

\begin{figure}[t]
	\centering
        \vspace{-3mm}
        \subfloat{
		\includegraphics[width=\linewidth]{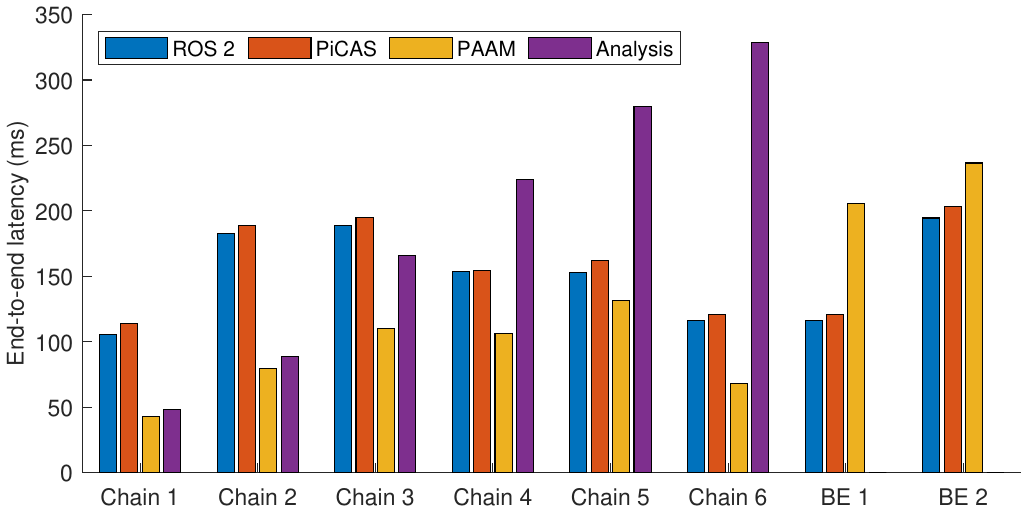}
	}
	\vspace{-1mm}
	\caption{Case study 1: Maximum observed chain latency}
	\label{fig:wc_case_study}
\end{figure}
\begin{figure}[t]
	\centering
	\subfloat{
		\includegraphics[width=\linewidth]{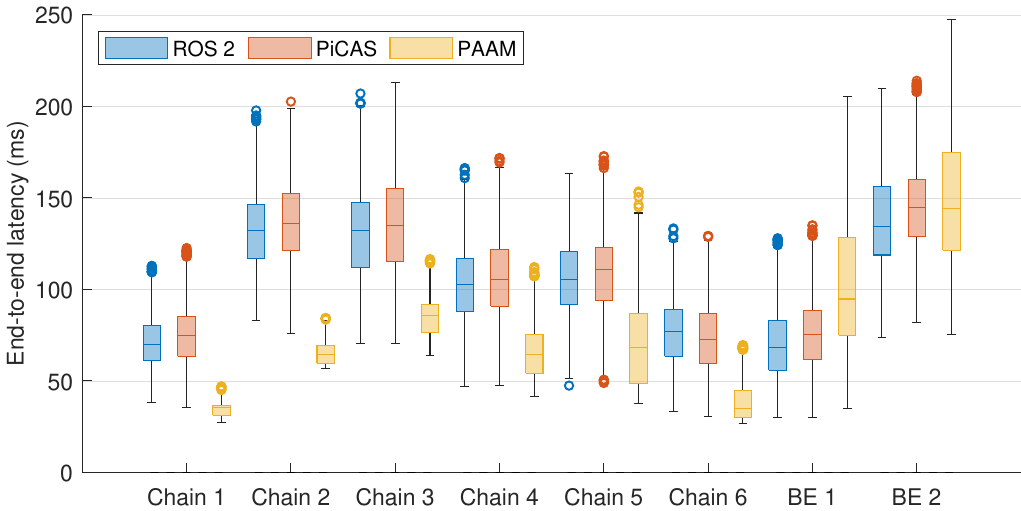} 
	}
	\vspace{-1mm}
	\caption{Case study 1: Chain latency distributions }
        \vspace{-1mm}
	\label{fig:ac_case_study}
\end{figure}

For comparison, we consider the following methods: 
\begin{itemize}
    \item \textbf{ROS~2}: Clients run on the default ROS 2 executors and use the direct invocation method for accelerator access.
    \item \textbf{PiCAS}~\cite{choi2021picas}: Clients execute as in ROS 2, but on the PiCAS executors with the direct invocation method.
    \item \textbf{PAAM}: Our proposed framework.
    \item \textbf{Analysis}: The worst-case end-to-end latency (response time) computed by our analysis given in Sec.~\ref{sec:analysis}. 
\end{itemize}

\smallskip\noindent\textbf{Results.} 
Fig.~\ref{fig:wc_case_study} compares the maximum chain latency observed from experiments under each method, along with the worst-case latency computed by our analysis. Fig.~\ref{fig:ac_case_study} reports the distributions of observed chain latencies under respective methods. 
We observed that the maximum observed chain latency under PAAM was always bounded by our theoretical analysis. This observation is juxtaposed with the unbounded response time of the highest-criticality chains without PAAM. In addition, the maximum observed latency of critical chains under PAAM is much shorter than ROS 2 and PiCAS (e.g., more than 50\% reduction for chain 1), which demonstrates the effectiveness of our proposed framework.

\subsection{Case Study 2: Autoware Reference System w/ GPU \& TPU}

\noindent\textbf{System Setup.}
To evaluate the benefit of PAAM in a more complex scenario, we used the Apex.AI's Autoware reference system~\cite{web:ros2refsys}. The system resembles the lidar-based perception pipeline of Autoware.Auto~\cite{web:autowareauto}, as illustrated in Fig.~\ref{fig:autoware_model}. We augmented this system by incorporating GPU and TPU segments into every callback, excluding the seven dedicated to the behavior planner to prevent accelerator overload. Recall that the Coral TPU does not allow for more than one context to exist at a time; so, to fairly evaluate the impact of PAAM on the reference system, we assigned a TPU inferencing workload to the single rear lidar points transformer callback. This setup permitted a straightforward comparison between the PAAM and direct invocation methods, with the latter requiring only one TPU context.
The reference system produces several Key Performance Indicators (KPIs):

\begin{itemize}
    \item {\bf Hot Path Latency}: The hot path, outlined in Fig.~\ref{fig:autoware_model}, is composed of the chain of callbacks from the lidar sensor to the collision estimator. This is the most critical chain in the system as it indicates the time to react to obstacles. 
    \item {\bf Behavior Planner Period}: The behavior planner is set to execute with a period of 100 ms. If the execution is blocked, the actual period may be prolonged. Lower variations in the measured period from the planner period are better.
    \item {\bf Hot Path Message Drops}: Dropped hot path messages show how many callbacks in the hot path missed their input messages in the current instance of the chain. A lesser amount of dropped messages indicates better performance. 
\end{itemize}

Unlike the first case study, we evaluate each method with both single-threaded (ST) and multi-threaded (MT) executors. 
\begin{itemize}
    \item \textbf{4$\times$ST:} We partitioned the reference system workload to four ST executors, each pinned to a separate CPU core. 
    \item \textbf{MT:} The system runs in one MT executor. To use the same number of cores as 4$\times$ST, we configured the number of threads to 4 and pinned the threads to to 4 cores.
\end{itemize}

\begin{figure}[t]
	\centering        
 	\subfloat[Hot Path Latency]{\label{fig:autoware_latency}            
		\includegraphics[width=.98\linewidth]{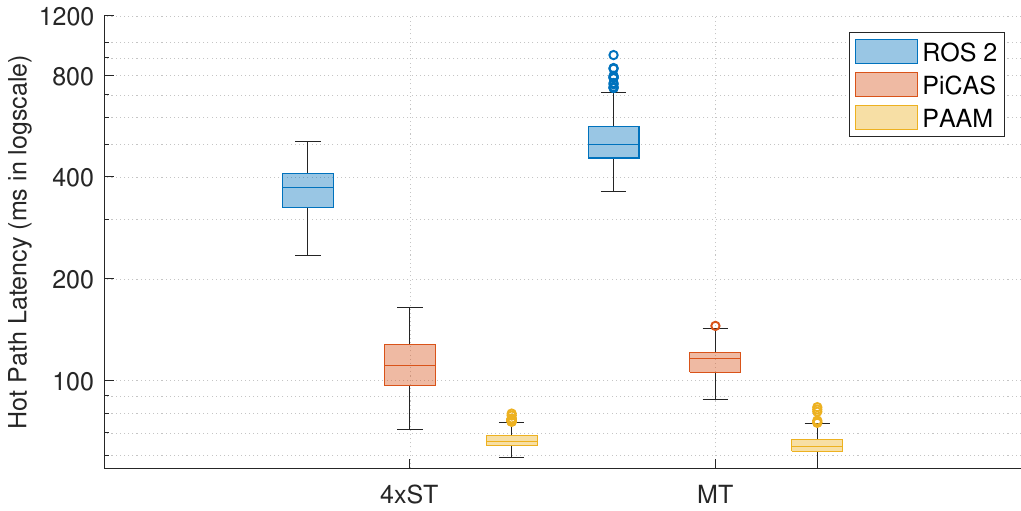}
	}\\\vspace{-3mm}
	\subfloat[Behavior Planner Period]{\label{fig:autoware_period}
		\includegraphics[width=.48\linewidth]{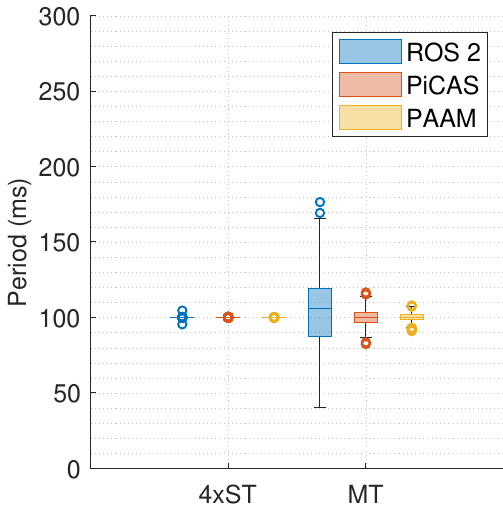}
	}        
        \subfloat[Hot Path Drops]{\label{fig:autoware_drops}
		\includegraphics[width=.48\linewidth]{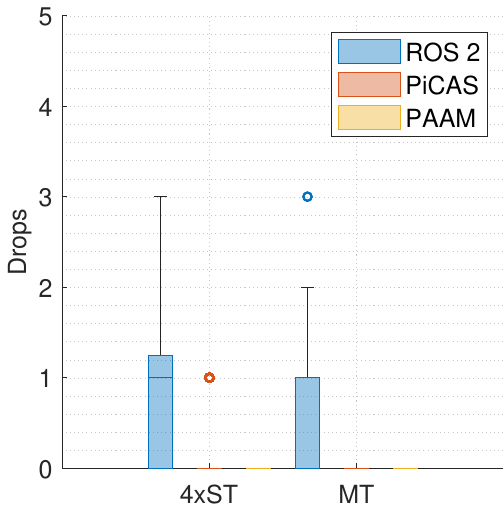}
	}
	\vspace{-1mm}
	\caption{Case Study 2: Autoware reference system results}
	\label{fig:autoware_results}
\end{figure}

\noindent\textbf{Results.} Fig.~\ref{fig:autoware_latency} depicts the end-to-end latency of the hot path across all system configurations. Overall, PAAM outperformed the rest, with PiCAS coming in second and ROS 2 trailing behind.
Focusing on the 4$\times$ST case, the priority-based chain scheduling of PiCAS achieves a 68\% reduction in the maximum hot path latency over ROS~2, decreasing from 510~ms to 163~ms. On top of this, PAAM further reduces the latency by 51\% compared to PiCAS, bringing it down to 79~ms. The substantial improvement by PAAM can be attributed to its ability to handle numerous low-criticality chains competing for accelerators in the reference system. The MT case follows the same trend, but interestingly, all three methods maintain poorer worst-case performance than in the ST$\times$4 case. This is because, with ST, the hot path encounters less blocking from low-criticality chains.


Figs.~\ref{fig:autoware_period} and \ref{fig:autoware_drops} show the observed distribution of behavior planner period and the number of dropped messages per hot path instance, respectively. PAAM achieves much lower variations from the preset 100 ms period, while simultaneously ensuring no message drops. From all these results, we conclude that PAAM can yield a significant improvement over the state-of-the-art, especially in a complex robotic environment.

\subsection{PAAM Overhead Analysis}\label{sec:ovh_brkdn}
	
\begin{figure}[t]
	\centering
	\subfloat{
		\includegraphics[width=\linewidth]{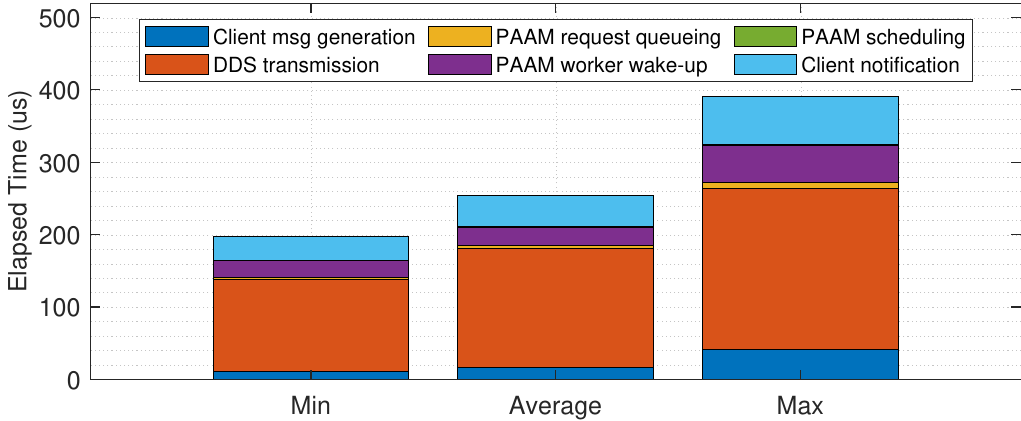}
	}
	\vspace{-2mm}
	\caption{Overhead breakdown}
        \vspace{-1mm}
	\label{fig:breakdown}
\end{figure}

\noindent\textbf{Overhead Breakdown.}
We sought to unpack the overhead associated with PAAM's servicing of client requests, in the context of best, average, and worst-case scenarios. As shown in Figure~\ref{fig:breakdown}, the overhead arising from PAAM's utilization spans from 198 $\mu$s in the optimal scenario to 391 $\mu$s in the most unfavorable one. 
A careful analysis reveals that a significant portion of the time is consumed by the DDS transport, sending a request message to the server in the control plane. The Iceoryx-enabled Cyclone DDS that the PAAM implementation uses is reported to have a latency of only a few $\mu$s for 1 KB payload~\cite{web:iceoryx_latency}~\cite{web:TSC-RMW-Reports}, which is much smaller than our control-plane message size (72 B). However, we found that its latency increases significantly when integrated into the ROS 2 ecosystem due to additional interface layers. This result further reinforces the rationale behind our design choice to have a separate data plane. Interestingly, we found that PAAM's request queueing, worker awakening, and scheduling, consume the least amount of time. 

\smallskip\noindent\textbf{GPU Preemption Delay with Multiple Buckets.}
Since our framework leverages a multi-bucket approach with preemptive execution for supported accelerators, we measured the GPU preemption delay that is incurred when executing a kernel on a high-priority stream while a low-priority stream is also executing a kernel. 
{To facilitate this, we created two prioritized streams within the same process context, one with higher priority than the other, allocated and assigned memory for two identical sets of kernels,
and asynchronously launched the first kernel on the lower priority stream first, then immediately launched the second kernel on the higher priority stream. 
We used the CUDA stream event timers to capture the execution time of both kernels and subsequently re-ran the identical kernel on the higher priority stream to establish a baseline runtime. We measured the preemption delay to be the difference between the isolated high-priority kernel’s execution time and the execution time of the high-priority kernel preempting the lower-priority kernel. We performed this benchmark with four separate types of kernels and reported the results of 50,000 iterations per kernel in Table~\ref{table:pd}. We observed that the maximum measured preemption delay was $129\mu s$. }

\begin{table}[t]
\centering
\vspace{-1mm}
\begin{tabular}{|l|l|l|l|l|}
\hline
               & MatMul & Reduction & VectorAdd & Histogram \\ \hline
Mean ($\mu s$)  & 41.21   & 39.55     & 22.78   & 14.95     \\ \hline
Max ($\mu s$)   & 116.48  & 129.18    & 77.85   & 72.92     \\ \hline
Stdev ($\mu s$) & 14.16   & 26.65     & 15.13   & 10.77     \\ \hline

\end{tabular}
\vspace{-1mm}
\caption{GPU bucket preemption cost}
\vspace{-1mm}
\label{table:pd}
\end{table}

\begin{figure}[t]
	\centering
        \vspace{-2mm}
	\subfloat{
		\includegraphics[width=\linewidth]{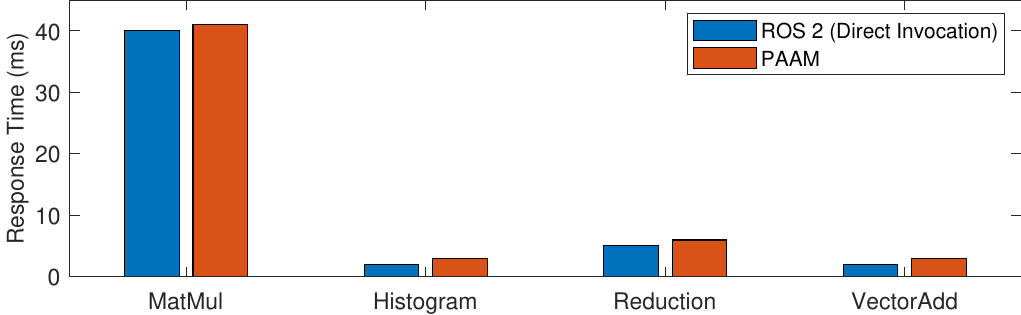}
	}
        \vspace{-2mm}
	\caption{Worst-case kernel execution time}
	\label{fig:microbench}
\end{figure}

\smallskip\noindent\textbf{Kernel Execution Time.}
To assess the impact of overhead in kernel execution, we report the execution time of several well-known GPU benchmark kernels when each runs alone in the system with and without PAAM. The results are depicted in Figure~\ref{fig:microbench}. The round-trip execution time of kernels with the PAAM server is only slightly longer compared to kernels that are directly invoked without the framework. We consider such a small increase in kernel response time acceptable given the huge benefit achievable by our framework. 

\subsection{Analytical Study} \label{rand_chain}
To explore the performance characteristics of our analysis, we conduct two experiments that examine the schedulability of chainsets based on varying parameters: the number of chains in a chainset and the accelerator-to-CPU utilization ratios. 

\begin{figure}[t]
	\centering
        \vspace{-7mm}
        \subfloat[\# of chains per chainset]{\label{fig:vc}\includegraphics[width=0.48\linewidth]{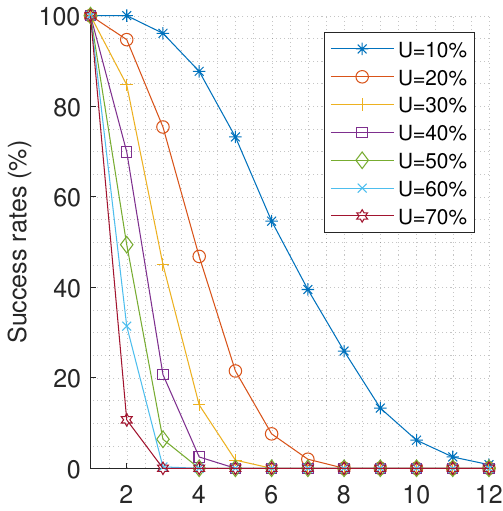}}     
  	\subfloat[Accelerator util. : CPU util.]{\label{fig:vr}\includegraphics[width=0.48\linewidth]{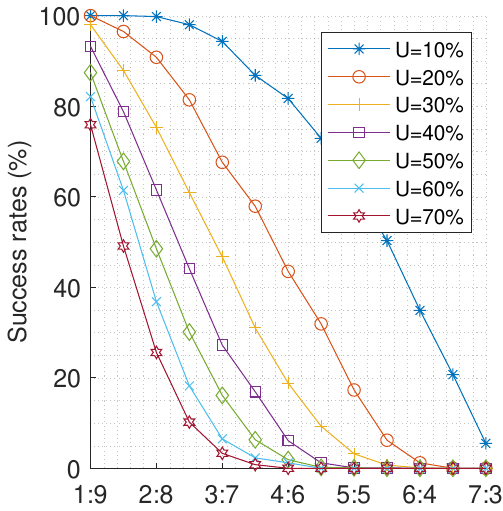}}   
	\vspace{-1mm}
        \caption{Schedulability of chainsets}
	\label{fig:vcr}
\end{figure}

For the first experiment, we established a fixed chain length of 4 callbacks per chain for a system with one accelerator. We then executed 1,000 schedulability tests for each distinct number of chains per chainset. The chains generated had random periodicity, but we ensured a strict 1:1 Accelerator:CPU utilization ratio per period, with the total utilization equally distributed amongst all the callbacks within the chain. Figure~\ref{fig:vc} shows the results, conducted with a varying number of chains per chainset. 

The second experiment was designed in a similar manner, but 
the ratio of accelerator utilization to CPU utilization per chain was varied from 1:9 to 7:3. The results of these schedulability tests are illustrated in Figure~\ref{fig:vr}. Interestingly, even when the utilization per chain remains constant, the schedulability of a chainset decreases as the accelerator-to-CPU utilization ratio escalates. This can be attributed to the contention on the singular accelerator that all chains and callbacks share. \par

\section{Related Work}\label{sec:related}

\subsection{Real-time Support For Robotic Applications}
There have been several prior studies conducted to enhance the analyzability and real-time performance of ROS. 
Casini et al.~\cite{casini2019response} modeled the default ROS 2 single-threaded executor and provided a response time analysis of a chain. They proposed using the Compositional Performance Analysis (CPA)~\cite{Henia2005} to find the end-to-end latency of a chain across executors. Tang et al.~\cite{tang2020response} also explored the real-time scheduling of chains on the default ROS 2 single-threaded executor and expanded upon \cite{casini2019response} to reduce analytical pessimism. 
Choi et el.~\cite{choi2021picas} proposed PiCAS to support priority-driven chain scheduling in ROS 2, enabling it to significantly outperform standard ROS 2 executors. 

Beyond executor analysis and accelerator support, there is also some work done to enhance other aspects of ROS. RobotCore~\cite{Robotcore} provides a vendor-agnostic interface to facilitate the use of heterogeneous accelerators in ROS 2 applications. TZC~\cite{TZC} leverages shared memory and partial serialization to reduce the performance overhead of message transport between processes. 

\subsection{Real-time Accelerator Support}
\label{rel_work_rtas}
Accelerator servers are not a new concept and many studies have been done to improve the real-time support of GPUs and other accelerators at the OS or system level. Kim et al.~\cite{kim2017server,Kim_JSA18} constructed a GPU server to manage real-time GPU requests at the system level. CARSS~\cite{carss2020} is a framework providing an interface for soft real-time workloads to utilize shared accelerator resources. The authors of \cite{Liang2015} and \cite{Tanasic2014} developed techniques for concurrent and preemptive GPU execution, but with no explicit consideration for real-time systems. Casini et al.~\cite{Casini_2022_het} proposed a framework for task partitioning of AD software on heterogeneous platforms with accelerators. While their framework offers response time bounds and is applicable to many robotic systems, it does not consider chain scheduling in the ROS 2 ecosystem, which is the focus of our work.

Recently, Li et al.~\cite{rosgm} presented ROSGM, a real-time GPU management framework for ROS 2. ROSGM is a node that can be added to an executor process, focusing on providing multiple GPU scheduling policies and arbitrating GPU requests from other nodes. This work, while solving partially the problem of direct invocation through the arbitration within an executor, does not consider end-to-end timing guarantees on processing chains. Our work addresses this problem, supports heterogeneous accelerators, allows multiple accelerator segments per callback, and solves fully the direct invocation problem by supporting multiple executor processes, which is common in AD systems such as Autoware~\cite{kato2018autoware}, with extensive consideration to mitigate inter-executor data exchanges.

\section{Conclusion}\label{sec:concl}

In this paper, we proposed PAAM, a coordinated and priority-driven framework for predictable accelerator access management in ROS 2. We demonstrated through our evaluation that PAAM brings about a significant improvement over the state-of-the-art and the default ROS 2 executors, with up to a 91\% decrease (from 917 ms to 83 ms) in the end-to-end response time of critical chains under realistic autonomous driving scenarios. PAAM also provides analytical bounds, allowing users to predict and test the timing behavior of their systems in the presence of heterogeneous accelerators.
With these, we believe that PAAM can serve as a promising solution for modern autonomous robotic systems with accelerators. 

\section*{Acknowledgment}
This work was sponsored by the National Science Foundation (NSF) grants 1943265, 1955650, and 2312395.


\bibliographystyle{IEEEtran}
\bibliography{draft/references}


\end{document}